\documentclass[letterpaper]{article} 
\usepackage{aaai24}
\usepackage{times}  
\usepackage{helvet}  
\usepackage{courier}  
\usepackage[hyphens]{url}  
\usepackage{graphicx} 
\usepackage{natbib}  
\usepackage{caption} 
\usepackage{algorithm}
\usepackage{algorithmic}
\usepackage{amssymb}
\usepackage{amsmath}
\usepackage{multirow}
\usepackage{booktabs}
\usepackage{color}  
\usepackage{svg}
\usepackage{newfloat}
\usepackage{listings}
\DeclareCaptionStyle{ruled}{labelfont=normalfont,labelsep=colon,strut=off} 
\floatstyle{ruled}
\newfloat{listing}{tb}{lst}{}
\floatname{listing}{Listing}
\usepackage{color}

\title{Unified Single-Stage Transformer Network for Efficient RGB-T Tracking}
\author{
    Jianqiang Xia\textsuperscript{\rm 1,\rm 3},
    DianXi Shi\textsuperscript{\rm 2,\rm 3},
    Ke Song\textsuperscript{\rm 5}, 
    Linna Song\textsuperscript{\rm 4},
    XiaoLei Wang\textsuperscript{\rm 2},
    Songchang Jin\textsuperscript{\rm 2},\\
    Li Zhou\textsuperscript{\rm 7}, Yu Cheng\textsuperscript{\rm 8}, Lei Jin\textsuperscript{\rm 11}, Zheng Zhu\textsuperscript{\rm 6}, Jianan Li\textsuperscript{\rm 9}, Gang Wang\textsuperscript{\rm 10}, Junliang Xing\textsuperscript{\rm 6}, Jian Zhao\textsuperscript{\rm 1}
}
\affiliations{
    \textsuperscript{\rm 1}National Innovation Institute of Defense Technology, 
    \textsuperscript{\rm 2}Intelligent Game and Decision Lab,\\ \textsuperscript{\rm 3}Tianjin Artificial Intelligence Innovation Center, 
    \textsuperscript{\rm 4}National University of Defense Technology,
    \textsuperscript{\rm 5}ShanDong University\\

    \textsuperscript{\rm 6}Tsinghua University, 
    \textsuperscript{\rm 7}Peking University, 
    \textsuperscript{\rm 8}National University of Singapore, \textsuperscript{\rm 9}Beijing Institute of Technology. \\ \textsuperscript{\rm 10}Beijing Institute of Basic Medical Sciences,
    \textsuperscript{\rm 11}Beijing University of Posts and Telecommunications
    
}

\begin{document}

\maketitle

\begin{abstract}
Most existing RGB-T tracking networks extract modality features in a separate manner, which lacks interaction and mutual guidance between modalities. This limits the network's ability to adapt to the diverse dual-modality appearances of targets and the dynamic relationships between the modalities. Additionally, the three-stage fusion tracking paradigm followed by these networks significantly restricts the tracking speed. To overcome these problems, we propose a unified single-stage Transformer RGB-T tracking network, namely USTrack, which unifies the above three stages into a single ViT (Vision Transformer) backbone with a dual embedding layer through self-attention mechanism. With this structure, the network can extract fusion features of the template and search region under the mutual interaction of modalities. Simultaneously, relation modeling is performed between these features, efficiently obtaining the search region fusion features with better target-background discriminability for prediction. Furthermore, we introduce a novel feature selection mechanism based on modality reliability to mitigate the influence of invalid modalities for prediction, further improving the tracking performance. Extensive experiments on three popular RGB-T tracking benchmarks demonstrate that our method achieves new state-of-the-art performance while maintaining the fastest inference speed 84.2FPS. In particular, MPR/MSR on the short-term and long-term subsets of VTUAV dataset increased by 11.1$\%$/11.7$\%$ and 11.3$\%$/9.7$\%$.
\end{abstract}

\section{Introduction}

\begin{figure}[t]
	\centering
	   \includegraphics[width=1.0\linewidth]{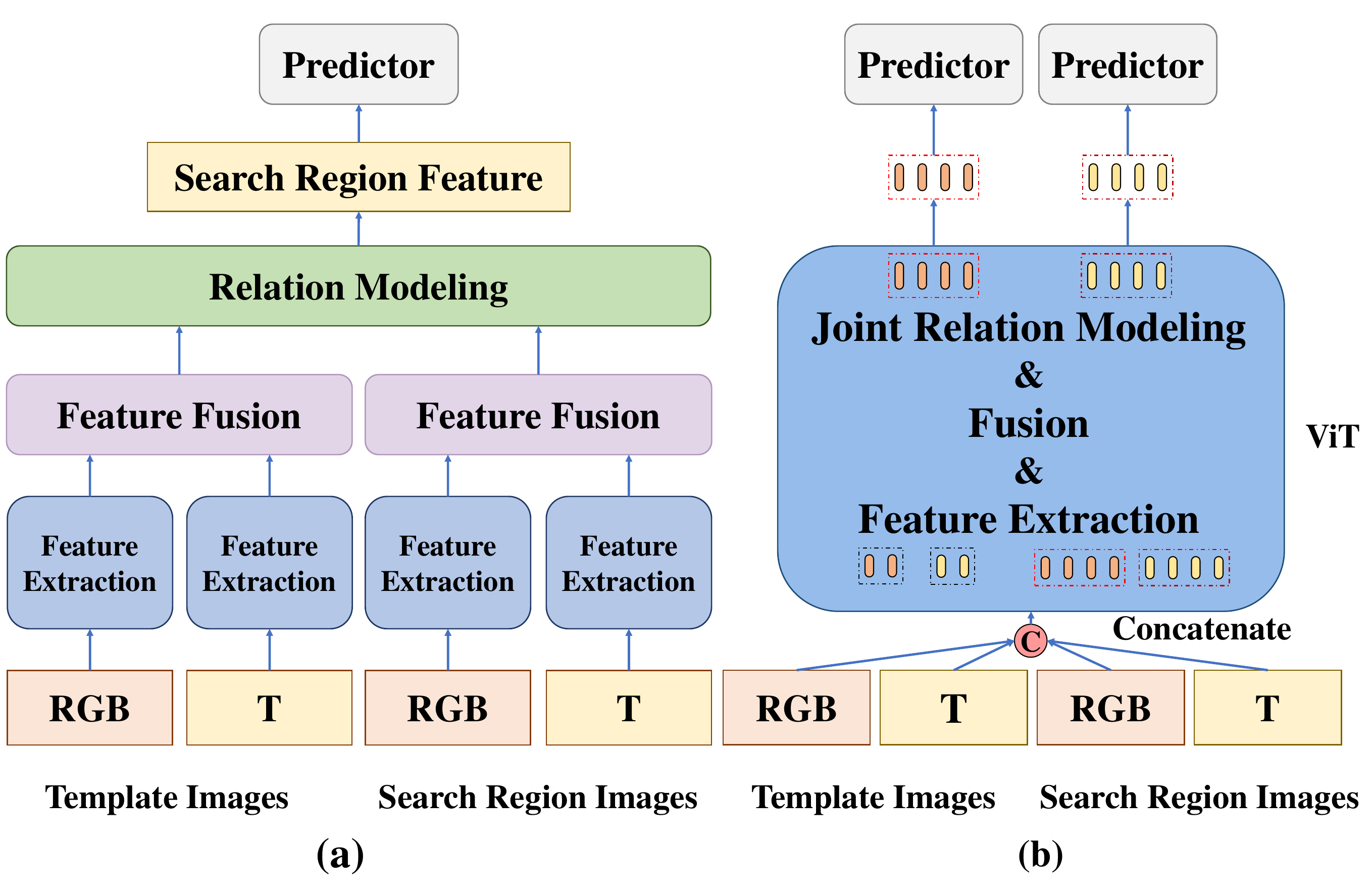}
	\caption{\small (a) Existing RGB-T tracking framework which adopts the RGB tracking network as basic architecture follows a three-stage tracking paradigm, extracting features by CNN or Transformer, fusing features of two modalities through additional customized fusion modules and relation modeling by online-training, cross-correlation, discriminative correlation or attention mechanism.  (b) Our RGB-T tracking framework performs joint feature extraction, fusion, and relation modeling by unifying the above three parts into a single ViT backbone. Best viewed in color.}\label{fig1}
    \vspace{-0.65cm}
\end{figure}

Visible-Thermal (RGB-T) tracking greatly expands the application scenarios of single object tracking (SOT) by using both RGB and thermal information, improving the tracking performance of SOT under challenging conditions such as illumination variation, occlusion, and extreme weather. Therefore, RGB-T tracking has become a research hotspot in recent years. Most RGB-T tracking network can be divided into three functional parts: feature extraction, feature fusion, and relation modeling between the fusion features of the template and the search region. Thanks to the rapid development of RGB tracking, existing RGB-T tracking networks directly adopt RGB tracking networks as the basic network architecture. They inherit the original manners of feature extraction and relation modeling, and then focus on the design of fusion modules. Their overall framework can be shown in Fig.~\ref{fig1}(a).

\begin{figure}[t]
	\centering
	\includegraphics[width=1.0\linewidth]{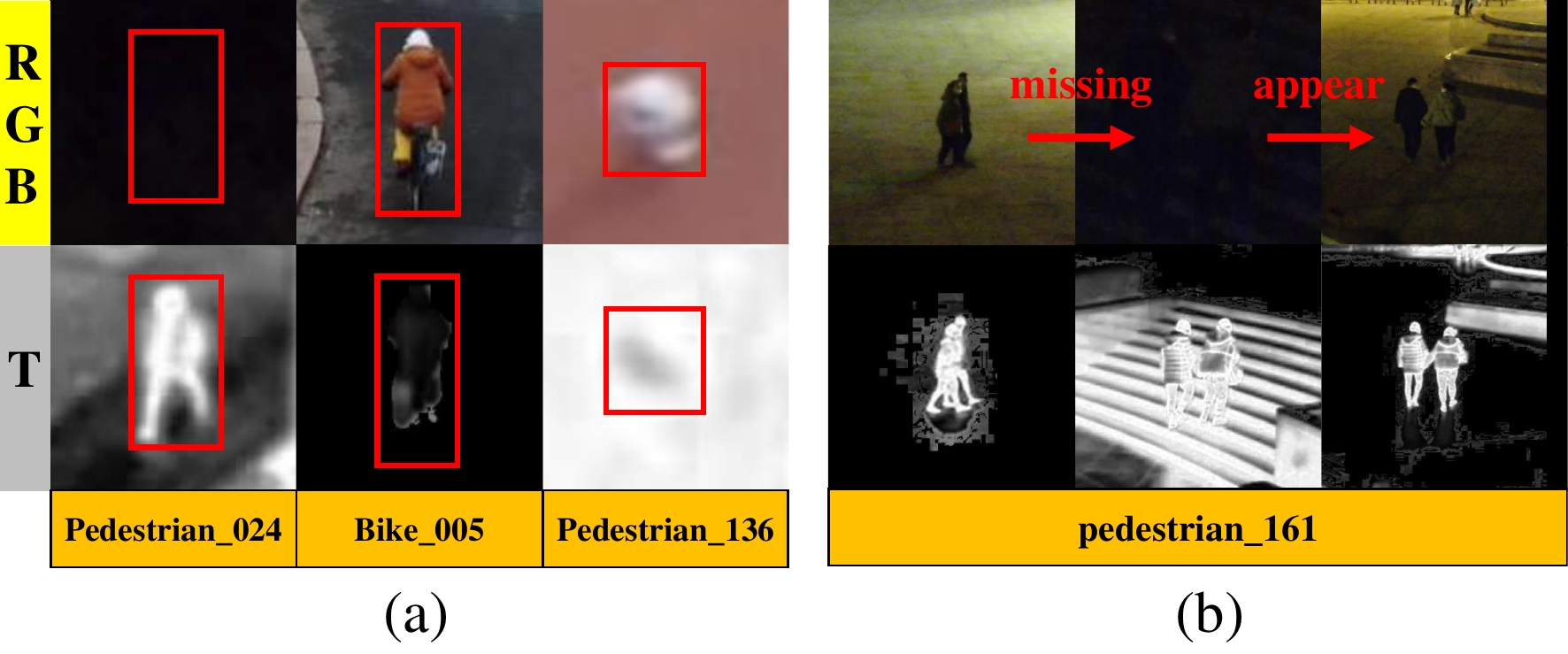}
	\caption{\small (a) Diverse dual-modality appearances of targets sampled in VTUAV \cite{HMFT-VTUAV} dataset. (b) During the dynamic tracking process, the state of the RGB appearance of the target is constantly changing. Best viewed in color.}\label{fig2}
 \vspace{-0.5cm}
\end{figure}

Most existing RGB-T tracking methods follow a three-stage fusion
tracking paradigm. they employ two subnetworks to extract RGB and thermal features separately from the template and search region. These features are then fused using a feature fusion module to obtain the template fusion features and the search region fusion features. Subsequently, a relation modeling operation between the fusion features of the template and search region is performed. After relation modeling, the processed search region fusion features are then utilized for prediction. However, the separate subnetworks lead to the lack of interaction between the two modalities during the feature extraction stage. As a consequence, the network can only extract regular features from each modality, rather than the dynamic features with an effective adjustment based on the state of modalities. However, as shown in Fig.~\ref{fig2}, such pattern is not fit to RGB-T tracking especially in complex environments, because different targets have diverse dual-modality appearances, and the appearances of both modalities can change continuously with the tracking environment. Temporary changing or missing appearances in the corresponding modality frequently happened due to the factors like occlusion, illumination variation, or thermal, which leads to the regions covered by the appearances of both modalities are not always consistent. In addition, three-stage fusion tracking paradigm is really difficult to balance the performance and speed.

We propose a unified single-stage Transformer RGB-T tracking network USTrack to solve the above problems. As shown in Fig.~\ref{fig1}(b), the core of USTrack is to unify feature extraction, feature fusion, and relation modeling into a single ViT \cite{ViT} backbone through the self-attention mechanism for simultaneous execution, efficiently obtaining search region fusion features used for prediction. Specifically, we first map the image patches from two modalities to appropriate latent spaces through a dual embedding layer to align the patterns and mitigate the impact of intrinsic heterogeneity for feature fusion. Within the attention layer of the ViT backbone, we perform the same operation on the template and search region, respectively. First, we concatenate the tokens from both modalities and then apply a self-attention mechanism to the concatenated features, directly extracting the template fusion features and search region fusion features. This process unifies the extraction and fusion of modality features, facilitating interaction between modalities during the feature extraction stage. The network can adaptively learn the semantic similarity between two modalities features based on attention weights, and use this similarity to model modality-sharing information. In the feature extraction stage, one modality can selectively acquire modality-specific information from another modality based on modality-sharing information, thereby guiding and adjusting the features to be extracted from itself. This enables the network to better adapt to diverse dual-modality appearances of targets and the dynamic relationships between the modalities. 

For the method of relation modeling between the template fusion features and the search region fusion features, inspired by the RGB tracking methods OSTrack  \cite{OSTrack} and SimTrack \cite{SimTrack}, we adopt performing the self-attention on the concatenated template and search region fusion features as our relation modeling manner. so that the network can enhance the target-background discriminability of the extracted search region fusion features under the guidance of two templates. To further improve the inference speed of the network without adding additional attention layers, we further unify the self-attention used for extracting fusion features with the self-attention used for relation modeling seamlessly. Through parallel execution, we significantly accelerate the inference speed.

In this way, we can obtain two search region fusion features based on different modalities for predicting the results. Unlike other transformer-based RGB-T tracking networks \cite{APFNet,MIRNet,TBSI} that directly concatenate two fusion features, we propose a feature selection mechanism based on modality reliability to reduce the impact of noise information of invalid modality on prediction. This mechanism can adaptively select the fusion feature more suitable for the tracking environment for prediction, further improving the tracking performance. To the best of our knowledge, USTrack is currently the first network to efficiently achieve RGB-T tracking without the use of any additional fusion modules. Our contributions are summarized as follows:

\begin{itemize}
    \item We propose a single-stage Transformer RGB-T tracking network USTrack which can extract fusion features of templates and search regions under the interaction of modalities, and simultaneously perform the relation modeling to further improve the tracking speed. 
    \item We propose a feature selection mechanism based on modality reliability, which can select appropriate fusion features from two fusion features based on different modalities according to the specific tracking environment to predict the results.
    \item USTrack exhibits state-of-the-art performance on benchmark GTOT \cite{GTOT}, RGBT234 \cite{RGBT234}, and VTUAV while maintaining the fastest inference speed at 84.2 FPS, Our source codes, pre-trained models, and online demos will be released upon acceptance.
\end{itemize}

\section{Related Work}

\noindent\textbf{RGB-T Tracking.}
In this part, we only summarize the Transformer-based RGB-T tracking methods. With the introduction of Transformer into RGB-T tracking, attention mechanism was first attempted for feature fusion. DRGCNet \cite{DRGCNet} and MIRNet \cite{MIRNet} use cross-attention to enhance discriminative features from one modality to another, and assign adaptive weights to features of two modalities through gating mechanism to filter redundant and noise information. APFNet \cite{APFNet} proposes an attribute-based progressive fusion network, which enhances the discriminative information specific to challenging attributes through cross-attention. However, the aforementioned Transformer-based RGB-T tracking methods are designed within a detection-based tracking framework\cite{MDNet}. On one hand, during the feature extraction stage, the modality features lack interaction due to the limited global context modeling capability of convolutional neural networks. On the other hand, although RGB-T tracking networks \cite{MIRNet} based on RT-MDNet \cite{RT-MDNet} have barely achieved real-time inference speed, they still follow a three-stage tracking paradigm, which first extracts modality features separately, then fuses features through various attention mechanism, and finally perform the relation modeling operation between the template and search region through online training and continuous fine-tuning, resulting in significant speed bottlenecks for these RGB-T tracking networks. 

The latest works TBSI \cite{TBSI} and ViPT \cite{ViPT} adopt the powerful and efficient RGB tracking network OSTrack \cite{OSTrack} as their base network architecture. However, they still design the fusion module as a separate component, which is inserted between two Transformer encoders to obtain the fused features of the template and search regions. The fused features are then fed into the encoders for joint feature extraction and relation modeling through self-attention mechanism. We summarize these latest RGB-T tracking methods as a two-stage RGB-T tracking network. It is worth noting that, unlike ViPT \cite{ViPT} that fixes the pre-training parameters of the backbone and only designs a simple fusion module to obtain the fusion features, TBSI \cite{TBSI} achieves feature fusion by inserting a complex cross-attention fusion module between Transformer encoders. It uses stacked extraction-fusion modules as the backbone of the network. This approach alleviates the lack of interaction between modalities during the feature extraction stage and significantly improves the performance. However, due to the addition of the extra complex cross-attention fusion module, it does not fully leverage the inference efficiency of OSTrack \cite{OSTrack}, resulting in barely achieving real-time performance. In order to achieve more efficient and concise interaction between modalities during the feature extraction stage, We attempt to unify feature extraction and feature fusion through the self-attention mechanism to directly extract fusion features of the template and search region, instead of designing complex additional feature fusion modules. 

\noindent\textbf{RGB Tracking.}
From the perspective of relation modeling methods, we provide a brief overview of the development of RGB tracking network frameworks which were adopted by existing RGB-T tracking methods as basic architecture. According to different tracking mechanism, early RGB tracking methods can be divided into two frameworks. One is detection-based tracking framework \cite{MDNet,RT-MDNet}, which models the relation between templates and search regions through online training and continuous fine-tuning, resulting in significant speed bottlenecks. The other is two-stream two-stage tracking framework, which performs the relation modeling through cross-correlation\cite{SiamFC}, discriminative correlation\cite{mfDimp}, or complex cross-attention\cite{TransT}. Methods based on this framework have fast inference speed. They extract the features of the template and the search region respectively, and then perform the relation modeling between them. However, two-stage tracking paradigm will limit the target-background discriminability of the search region features. RGB-T tracking networks using these tracking frameworks as the basic network architecture also face the same problems. 

To overcome the problems faced by early tracking frameworks, SimTrack \cite{SimTrack} and OSTrack \cite{OSTrack} perform relation modeling by directly applying self-attention on the concatenated templates and search region features, thereby enhancing the target-background discriminability of search region features under the guidance of templates. Inspired by them,  we attempt to directly perform self-attention on the concatenated template and search region fusion features as our relation modeling manner, which allows the network to extract the search region fusion features under the guidance of the two templates, obtaining the search region fusion features with better target-background discriminability. With the scalability  of self-attention mechanisms, we further unify the self-attention used for fusion feature extraction with the self-attention used for relation modeling for parallel execution, thereby significantly improving the inference speed of the RGB-T tracking network.

\section{Unified Single-Stage RGB-T Tracking}

\begin{figure*}[htbp]
	\centering
	\includegraphics[width=1.0\linewidth]{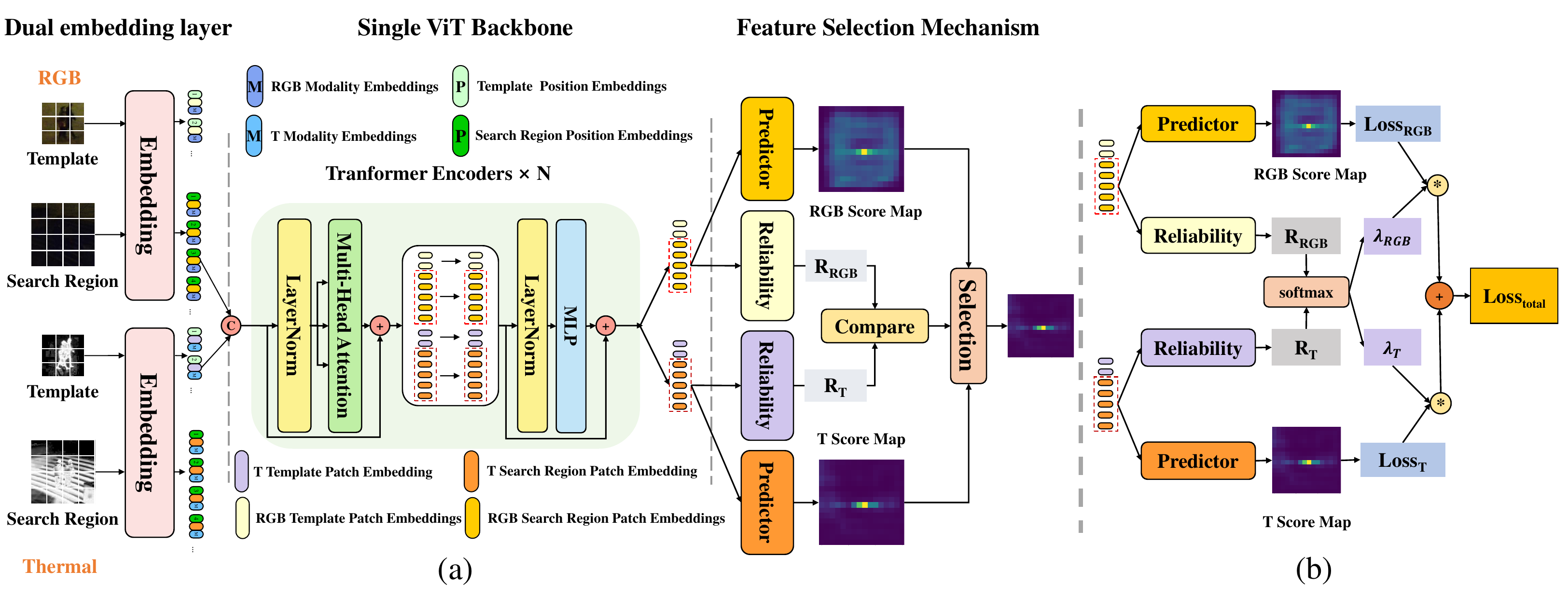}
	\caption{\small (a) The overall architecture of USTrack. The template and search region are split, flattened, and linear projected through the dual embedding layer. Image embeddings are then concatenated and fed into Transformer encoder layers for joint feature extraction, fusion and relation modeling. The feature selection mechanism is responsible for selecting fusion features with higher reliability for result prediction. (b) The training of the feature selection mechanism based on modality reliability. Best viewed in color.}\label{fig3}
 \vspace{-0.4cm}
\end{figure*}

\noindent\textbf{Overview.} As shown in Fig.~\ref{fig3}, the overall architecture of USTrack consists of three components: a dual embedding layer, a single ViT backbone and the dual prediction heads with a feature selection mechanism based on modality reliability. Self-attention is based on similarity to obtain global information, the inherent heterogeneity between modalities may limit the network's ability to model modality-sharing information through attention weights, thereby affecting the subsequent fusion process. Therefore, we use two learnable embedding layers to map inputs belonging to different modalities into a latent space that is conducive to fusion. We choose ViT as our backbone network, utilizing its self-attention layers to simultaneously perform feature extraction, feature fusion, and relation modeling between fusion features from the template and the search region, obtaining search region fusion features that contain relation information for prediction. Considering that a single modality may not be suitable for all tracking scenarios, such as darkness, occlusion and thermal crossover, where the corresponding modality has already lost all information about the target's appearance and generates a significant amount of noise information. Feature selection mechanism based on modality reliability helps the network select search region fusion features generated by the modality that is more suitable for the current tracking scenario, reducing the influence of noise caused by ineffective modalities on the prediction results.

\subsection{Dual Embedding Layer}

The input of USTrack is a pair of target template images and a pair of search region images, consisting of a total of four images, namely, the RGB template image $\small{\boldsymbol{z}_{image}^{rgb} \in \mathbb{R}^{H_z \times W_z \times 3}}$, the RGB search region image $\small{\boldsymbol{x}_{image}^{rgb} \in \mathbb{R}^{H_x \times W_x \times 3}}$, the thermal template image ${\boldsymbol{z}_{image}^{t} \in \mathbb{R}^{H_z \times W_z \times 3}}$ and the thermal search region image $\small{\boldsymbol{x}_{image}^{t} \in \mathbb{R}^{H_x \times W_x \times 3}}$. They are first split and flattened into sequences of patches $\small{\boldsymbol{z}_{rgb},\boldsymbol{z}_t\in\mathbb{R}^{N_z\times(3P^2)}}$ and $\small{\boldsymbol{x}_{rgb},\boldsymbol{x}_t\in\mathbb{R}^{N_x\times(3P^2)}}$, where $\small{P\times P}$ is the resolution of each patch, and $\small{N_z=\frac{H_zW_z}{P^2}}$, $\small{N_x=\frac{H_xW_x}{P^2}}$ are the number of patches of template and search region respectively. Then, two trainable linear projection layers with parameters $\small{\boldsymbol{E}_{rgb}\in\mathbb{R}^{(3P^2)\times D}}$ and $\small{\boldsymbol{E}_{t}\in\mathbb{R}^{(3P^2)\times D}}$ are used to project $\small{\boldsymbol{z}_{rgb}}$, $\small{\boldsymbol{x}_{rgb}}$ and $\small{\boldsymbol{z}_t}$, $\small{\boldsymbol{x}_t}$ into $\small{D}$ dimension latent space. The output of this projection are four patch embeddings $\small{\hat{\boldsymbol{z}}_{rgb}}$, $\small{\hat{\boldsymbol{x}}_{rgb}}$ and $\small{\hat{\boldsymbol{z}}_t}$, $\small{\hat{\boldsymbol{x}}_t}$. Learnable $\small{1D}$ position embeddings $\small{\boldsymbol{P}_z}$ and $\small{\boldsymbol{P}_x}$ are added to the template patch embeddings $\small{\hat{\boldsymbol{z}}_{rgb}}$, $\small{\hat{\boldsymbol{z}}_t}$ and search region patch embeddings $\small{\hat{\boldsymbol{x}}_{rgb}}$, $\small{\hat{\boldsymbol{x}}_t}$ separately, and Learnable $\small{1D}$ modality embeddings $\small{\boldsymbol{M}_{rgb}}$ and $\small{\boldsymbol{M}_t}$ are added to the RGB patch embeddings $\small{\hat{\boldsymbol{z}}_{rgb}}$, $\small{\hat{\boldsymbol{x}}_{rgb}}$ and thermal patch embeddings $\small{\hat{\boldsymbol{z}}_t}$, $\small{\hat{\boldsymbol{x}}_t}$ separately. The patch embeddings after adding position and modality embeddings are final features called token embeddings. The above operations can be represented as follows:
\begin{small}
\begin{equation}
\hat{\boldsymbol{z}}_{rgb} =\left[\boldsymbol{z}_{rgb}^1 \boldsymbol{E}_{rgb};\boldsymbol{z}_{rgb}^2\boldsymbol{E}_{rgb};...;\boldsymbol{z}_{rgb}^{N_z}\boldsymbol{E}_{rgb}\right]+\boldsymbol{P}_z+\boldsymbol{M}_{rgb},
\end{equation}
\end{small}
\begin{small}
\begin{equation}
\hat{\boldsymbol{z}}_t=\left[\boldsymbol{z}_t^1 \boldsymbol{E}_t;\boldsymbol{z}_t^2 \boldsymbol{E}_t;...;\boldsymbol{z}_t^{N_z} \boldsymbol{E}_t\right]+\boldsymbol{P}_z+\boldsymbol{M}_t,
\end{equation}
\end{small}
\begin{small}
\begin{equation}
\hat{\boldsymbol{x}}_{rgb} =[\boldsymbol{x}_{rgb}^1 \boldsymbol{E}_{rgb};\boldsymbol{x}_{rgb}^2 \boldsymbol{E}_{rgb};...;\boldsymbol{x}_{rgb}^{N_x} \boldsymbol{E}_{rgb}]+ \boldsymbol{P}_x+ \boldsymbol{M}_{rgb},
\end{equation}
\end{small}
\begin{small}
\begin{equation}
\hat{\boldsymbol{x}}_t=[\boldsymbol{x}_t^1 \boldsymbol{E}_t;\boldsymbol{x}_t^2 \boldsymbol{E}_t;...;\boldsymbol{x}_t^{N_x} \boldsymbol{E}_t]+ \boldsymbol{P}_x+ \boldsymbol{M}_t.
\end{equation}
\end{small}

After passing the dual embedding layer, RGB template token embeddings $\small{\hat{\boldsymbol{z}}_{rgb}}$, thermal template token embeddings $\small{\hat{\boldsymbol{z}}_t}$, RGB search region token embeddings $\small{\hat{\boldsymbol{x}}_{rgb}}$ and thermal search region token embeddings $\small{\hat{\boldsymbol{x}}_t}$ will be input into the backbone for subsequent processing.

\subsection{Single ViT Backbone}

The self-attention mechanism is the core component of the ViT, and it is also the key to performing joint feature extraction, feature fusion and relation modeling in a single ViT backbone. From the perspective of the self-attention mechanism, we take the RGB search region token embeddings $\small{\hat{\boldsymbol{x}}_{rgb}}$ as an example to further analyze the intrinsic reasons why the proposed network is able to realize simultaneous feature extraction, feature fusion and relation modeling.

In the attention layer, the token sequences $\small{\hat{\boldsymbol{x}}_{rgb}}$, $\small{\hat{\boldsymbol{x}}_t}$, $\small{\hat{\boldsymbol{z}}_{rgb}}$, $\small{\hat{\boldsymbol{z}}_t}$ from dual embedding layers are concatenated as $\small{\boldsymbol{H}=\begin{bmatrix}\widehat{\boldsymbol{x}}_{rgb};\widehat{\boldsymbol{x}}_t;\widehat{\boldsymbol{z}}_{rgb};\widehat{\boldsymbol{z}}_t\end{bmatrix}\in\mathbb{R}^{(2N_x+2N_t)\times D}}$. Then Self-attention operation is performed on $\small{\boldsymbol{H}}$ as follows: 
\begin{small}
\begin{equation}
\boldsymbol{M}=\boldsymbol{A}\cdot\boldsymbol{V}=softmax\left(\frac{\boldsymbol{Q} \boldsymbol{K}^T}{\sqrt{d_k}}\right)\cdot \boldsymbol{V},
\end{equation}
\end{small}
\begin{small}
\begin{equation}
\boldsymbol{Q} \boldsymbol{K}^T=[\boldsymbol{Q}^x_{rgb};\boldsymbol{Q}^x_t;\boldsymbol{Q}^z_{rgb};\boldsymbol{Q}^z_t][\boldsymbol{K}^x_{rgb};\boldsymbol{K}^x_t;\boldsymbol{K}^z_{rgb};\boldsymbol{K}^z_t]^T, \label{EQ6}
\end{equation}
\end{small}
\begin{small}
\begin{equation}
\boldsymbol{V}=\left[\boldsymbol{V}^x_{rgb};\boldsymbol{V}^x_t;\boldsymbol{V}^z_{rgb};\boldsymbol{V}^z_t\right],
\end{equation}
\end{small}where $\small{\boldsymbol{M}}$ is the output of self-attention operation. $\small{\boldsymbol{A}}$ is the attention weight. $\small{\boldsymbol{Q}}$, $\small{\boldsymbol{K}}$, and $\small{\boldsymbol{V}}$ are query, key and value matrices separately. The superscripts $\small{z}$ and $\small{x}$ denote matrix items belonging to the template and search region. The subscripts $\small{rgb}$ and $\small{t}$ denote matrix items belonging to the RGB modality and thermal modality. The calculation of attention weights in Eq.~(\ref{EQ6}) can be expanded to follows:
\begin{small}
\begin{equation}
    \begin{aligned}
    \boldsymbol{Q} \boldsymbol{K}^T &= [\boldsymbol{Q}^x_{rgb} {\boldsymbol{K}^x_{rgb}}^T,\boldsymbol{Q}^x_{rgb}  {\boldsymbol{K}^x_t}^T,\boldsymbol{Q}^x_{rgb} {\boldsymbol{K}^z_{rgb}}^T,\boldsymbol{Q}^x_{rgb} {\boldsymbol{K}^z_t}^T;...]  \\ 
    &=[\boldsymbol{W}^{x_{rgb}}_{x_{rgb}},\boldsymbol{W}^{x_{rgb}}_{x_t},\boldsymbol{W}^{x_{rgb}}_{z_{rgb}},\boldsymbol{W}^{x_{rgb}}_{z_t};...],
    \end{aligned}\label{EQ8}
\end{equation}
\end{small}where the left part of Eq.~(\ref{EQ8}) represents the calculation and the output of attention weights between the RGB search region tokens and the other inputs. the output of self-attention operation  can be further written as follows:

\begin{small}
\begin{equation}
	\begin{aligned}
            \boldsymbol{M} =\left[{\boldsymbol{W}^{x_{rgb}}_{x_{rgb}} \boldsymbol{V}^x_{rgb}+ \boldsymbol{W}^{x_{rgb}}_{x_t} \boldsymbol{V}^x_t}\right. \\ 
            \left.{+\boldsymbol{W}^{x_{rgb}}_{z_{rgb}} \boldsymbol{V}^z_t + \boldsymbol{W}^{x_{rgb}}_{z_t} \boldsymbol{V}^z_t;...}\right],
        \end{aligned}\label{Eq9}
\end{equation}
\end{small}where the left part of Eq.~(\ref{Eq9}) is the output corresponding to the RGB search region tokens after the self-attention operation. $\small{\boldsymbol{W}^{x_{rgb}}_{x_{rgb}} \boldsymbol{V}^x_{rgb}}$ is responsible for aggregating the RGB search region image feature (RGB modality feature extraction). $\small{\boldsymbol{W}^{x_{rgb}}_{x_t}\boldsymbol{V}^x_t}$ is responsible for aggregating the thermal modality-specific information based on semantic similarity between two modalities features (feature fusion and modality features interaction). The attention weights can intuitively measure the semantic similarity between modalities. Network can model modality-sharing information based on this similarity. The aggregation of complementary information enables the network to promptly adjust the subsequent extraction of features in RGB search region image. $\small{\boldsymbol{W}^{x_{rgb}}_{z_{rgb}} \boldsymbol{V}^z_t}$ is responsible for aggregating RGB template image feature to further obtain the relation information between the RGB template and the RGB search region (relation modeling based on modality-specific information). $\small{\boldsymbol{W}^{x_{rgb}}_{z_t} \boldsymbol{V}^z_t}$ is responsible for aggregating thermal template image feature to further obtain the relation information between the thermal template and the RGB search region (relation modeling based on modality-sharing information). The RGB search region fusion features, which contains relation information, can be used for prediction. Therefore, with the global perception ability of the self-attention, we seamlessly unify feature extraction, feature fusion, and relation modeling into a single ViT backbone. The network can directly extract fusion features of the template and search region under the mutual interaction of modalities, and simultaneously performs relation modeling between fusion features of the template and search region. This alleviates the lack of interaction and guidance between modalities during the feature extraction stage, as well as the problem of additional fusion modules significantly affecting the inference speed of the RGB-T tracking network. Additionally, by inheriting the advantages of relation modeling which is performed by the self-attention, the network can extract more target-specific search region fusion features for prediction under the guidance of two templates.

\subsection{Feature Selection Mechanism}
After passing the ViT backbone, two search region fusion features can be obtained for final prediction: Thermal-assisted RGB fusion features based on the RGB search region image, and RGB-assisted thermal fusion features based on the thermal search region image. Both fusion features contain the fusion information of modalities and the relation information between the template and the search region, which can be directly used for target position prediction. To fully exploit the modality-sharing information and the modality-specific information of the modality that is more suitable for the current scene. As shown in Fig.~\ref{fig3}(b), during the training phase, we equip each fusion feature with a prediction module and a reliability evaluation head. We set the same loss for each prediction head and let each reliability evaluation module output a weight as the adaptive reliabilty for the loss of the corresponding prediction head. Then, they are combined into a final total loss function for training. This method allows the modality that is not suitable for the current scene to produce inferior results, and then guides the modality reliability evaluation module to assign smaller weights to its loss by minimizing the overall loss function. Conversely, for fusion features that are more suitable for the current scene, it assigns larger weights. During the testing phase, the network will simultaneously output two results and evaluate the reliability of both modalities. Based on the reliability $\small{R_{RGB}}$ and $\small{R_T}$, we select the predicted results with higher reliability scores as the final output.

\subsection{Dual Prediction Heads and Loss}
We adopt the prediction head of OSTrack \cite{OSTrack} directly as our prediction head. The detailed information and corresponding settings can be found in OSTrack. The loss corresponding to the two prediction heads are set as follows:
\begin{equation}
\small{\mathcal{L}_{RGB}=\mathcal{L}_{cls_{RGB}}+\lambda_{giou}\mathcal{L}_{giou_{RGB}}+\lambda_{L_1}\mathcal{L}_{1_{RGB}}},
\end{equation}
\begin{equation}
\small{\mathcal{L}_T=\mathcal{L}_{cls_T}+\lambda_{giou}\mathcal{L}_{giou_T}+\lambda_{L_1}\mathcal{L}_{1_T}},
\end{equation}where $\small{\mathcal{L}_{RGB}}$ and $\small{\mathcal{L}_T}$ are the overall loss function for each prediction head, $\small{\mathcal{L}_{cls_{RGB}}}$ and $\small{\mathcal{L}_{cls_T}}$ are the weighted focal loss for classification, $\small{\mathcal{L}_{1_{RGB}}}$ and $\small{\mathcal{L}_{1_T}}$ are the $\small{\mathcal{L}_1}$ loss, $\small{\mathcal{L}_{giou_{RGB}}}$ and $\small{\mathcal{L}_{giou_T}}$ are the generalized IoU loss, and $\small{\lambda_{giou}}$ and $\small{\lambda_{L_1}}$ are the regularization parameters. On the basis, a modality reliability evaluation module is added to each search region fusion features. The evaluation module is a fully convolutional neural network, which consists of several stacked Conv-BN-ReLU layers. Two modality reliability evaluation modules will output the  reliability scores corresponding to two search region fusion features $\small{R_{RGB}}$, $\small{R_T\in\mathbb{R}}$ respectively. In order to prevent the model from directly making the weight $\small{R_{RGB}}$ and $\small{R_T}$ zero to minimize the overall loss during the training process, we softmax the reliability scores to obtains the adaptive weight $\small{\lambda_{RGB}}$ and $\small{\lambda_T}$ and overall loss as follows: 
\begin{equation}
	\small{\lambda_{RGB},\lambda_T=softmax(R_{RGB},R_T)},
\end{equation}
\begin{equation}
\small{\mathcal{L}_{total}=\lambda_{RGB}\mathcal{L}_{RGB}+\lambda_T\mathcal{L}_T},
\end{equation}where $\small{\lambda_{RGB}}$ and $\small{\lambda_T}$ are used as the adaptive weights of the loss of the two prediction heads, and the two losses are combined together as the overall loss to train the model.

\section{Experiment}
\subsection{Experiment Settings}

We compare our method with previous state-of-the-art RGB-T tracking methods on three benchmarks including VTUAV, RGBT234, and GTOT. GTOT and RGBT234 use success rate SR and precision rate PR as evaluation metrics. To mitigate small alignment errors, VTUAV use Maximum Precision Rate MPR and Maximum Success Rate MSR as evaluation metrics. SR measures the ratio of tracked frames, determined by the Interaction-over-Union (IoU) between tracking result and ground truth. With different overlap thresholds, a success plot (SP) can be obtained, and SR is calculated as the area under curve of SP. MSR adopts the maximum overlap in frame level as the final score. PR measures the percentage of frames whose distance between the predicted position and the ground truth is less than a certain threshold $\tau$. Similar to MSR, MPR adopt the smaller center distance as the final score. $\tau$ is set to 20 in our experiment.

Our model is implemented based on Python 3.8, PyTorch 2.0.0. All experiments are conducted on one NVIDIA RTX3090 GPU. We adopt AdamW as the optimizer with 1e-4 weight decay. The learning rate is set as 4e-5 for the backbone and 4e-4 for other parameters. The search regions are resized to 256×256 and templates are resized to 128×128. Each batch size is set to 24, and each epoch contains 30k image pairs. In order to fairly compare our method with other SOTA methods and demonstrate its effectiveness, we aligned our experimental conditions with other methods. We pretrained the network on RGB tracking datasets such as COCO \cite{COCO}, LaSOT \cite{LaSOT}, GOT-10k \cite{GOT-10k}, and TrackingNet \cite{Trackingnet}. When testing on the GTOT and RGBT234, we only used LasHeR as the training set. When testing on the short-term and long-term testing sets of VTUAV, we only use the training set of VTUAV for training.

\subsection{Comparison with State-of-the-art Methods}

We test our network USTrack on three popular RGB-T tracking benchmarks, comparing performance and speed with the SOTA trackers, such as FSRPN,  mfDimp, DAFNet , DAPNet, MANet, CAT, CMPP, JMMAC, MANet++, ADRNet, SiamCDA, M5LNet, TFNet, DMCNet, MFGNet, APFNet, HMFT, MIRNet, ECMD, ViPT and TBSI, to validate the effectiveness of our method. The test results on three datasets show that our method has achieved significant improvements in both performance and inference speed.

\begin{figure}[t]
	\centering
	\includegraphics[width=1.0\linewidth]{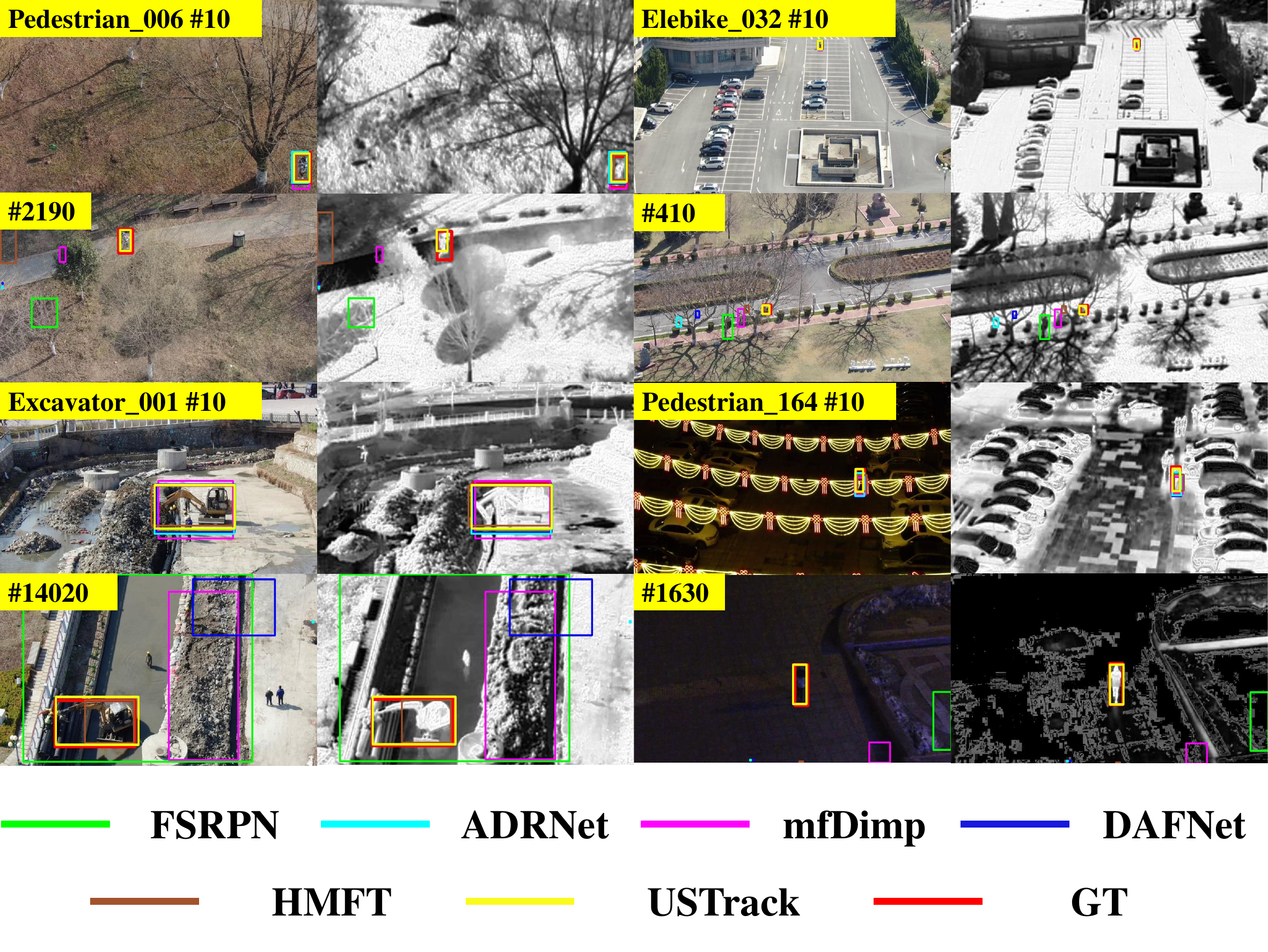}
	\caption{\small Visualization between our method and other RGB-T trackers on four representative sequences  which include multiple challenge attributes from VTUAV dataset. Best viewed in color.}\label{fig4}
 \vspace{-0.5cm}
\end{figure}

\noindent\textbf{Evaluation on VTUAV Dataset.}
VTUAV dataset is the latest and largest RGB-T tracking dataset, which is currently the only dataset that provides a test for long-term tracking performance of RGB-T tracking methods. Long-term sequences can effectively demonstrate the differences in the target appearances of different modalities, and the persistent changing relationships of two appearances state during the tracking process. As shown in Tab.~\ref{tab1} and Fig.~\ref{fig4}, despite USTrack being a short-term tracking network with no template updating or local-to-global strategies for long-term tracking, we still conduct tests on the short-term and long-term subsets of VTUAV to verify the tracking performance of USTrack. The results were very satisfactory. Compared to the best performing method HMFT and HMFT-LT, we achieved 11.1$\%$/11.7$\%$ and 11.3$\%$/9.7$\%$ increases in MPR/MSR on the VTUAV short-term dataset and VTUAV long-term dataset, respectively. Our speed was also 2.78 times and 10.4 times faster than the SOTA method HMFT and HMFT-LT \cite{HMFT-VTUAV}, significantly surpassing the baseline methods of the benchmark dataset. 

It is worth mentioning that, to our knowledge, HMFT-LT is currently the only long-term RGB-T tracking method that utilizes the local-to-global tracking strategy. We have achieved significantly better tracking performance on long-term datasets compared to HMFT-LT at ten times the speed of HMFT-LT. Furthermore, we have achieved the best results on all subsets of challenge attributes in VTUAV, especially those that will lead to continuous changes in the state of modality appearance, such as occlusion, extreme lighting, \textit{etc}. Detailed test results are provided in the supplementary materials. Above results demonstrate the excellent adaptability of our method to the diverse dual-modality appearances of targets and the dynamic relationship between modalities, fully demonstrating the effectiveness and efficiency of USTrack, which can directly extract dynamic fusion features of templates and search regions under the interaction of modalities, and simultaneously perform the relation modeling to further improve the tracking speed. 

\begin{table*}[t]\small
	\begin{center}
		\begin{tabular}{c|c|c|c|c|c|c|c|c|c|c}
			\toprule
			\multirow{2}{*}{Method} & \multirow{2}{*}{Pub} & \multicolumn{2}{|c|}{GTOT} & \multicolumn{2}{|c|}{RGBT234} & \multicolumn{2}{|c|}{VTUAV-short} & \multicolumn{2}{|c|}{VTUAV-long} & \multirow{1}{*}{Speed} \\ 
			 & & PR & SR & PR & SR & MPR & MSR & MPR & MSR & FPS\\
			\midrule
			FSRPN \cite{FSRPN} & ICCVW'19 &89.0&69.5&71.9&52.5&65.3&54.4&36.6&31.4&36.8  \\
			mfDimp \cite{mfDimp} & ICCVW'19 &83.6&69.7&84.6&59.1&67.3&55.4&31.5&27.2&34.6  \\
			DAFNet \cite{DAFNet} & ICCVW'19 &89.1&71.6&79.6&54.4&62.0&45.8&25.3&18.8&20.5  \\
			
			DAPNet \cite{DAPNet} & ACM MM'19 &88.2&70.7&76.6&53.7&-&-&-&-&-  \\
                MANet \cite{MANet} & TIP'20 &89.4&72.4&77.7&53.9&-&-&-&-&2.1  \\
                CAT \cite{CAT} & ECCV'20 &88.9&71.7&80.4&56.1&-&-&-&-&-  \\
                CMPP \cite{CMPP} & CVPR'20 &92.6&73.8&82.3&57.5&-&-&-&-&-  \\
                
                JMMAC \cite{JMMAC} & TIP'21   &90.2&73.2&79.0&57.3&-&-&-&-&-  \\
                MANet++ \cite{MANet++} & TIP'21 &88.2&70.7&79.5&55.9&-&-&-&-&25.4  \\
                ADRNet \cite{ADRNet} & IJCV'21 &90.4&73.9&80.7&57.1&62.2&46.6&23.5&17.5&25.0  \\
                SiamCDA \cite{SiamCDA} & TCSVT'21 &87.7&73.2&79.5&54.2&-&-&-&-&24.0  \\
                
                M5LNet \cite{M5LNet} & TIP'22 &89.6&71.0&79.5&54.2&-&-&-&-&9.0  \\
                TFNet \cite{TFNet} & TCSVT'22 &88.6&72.9&80.6&56.0&-&-&-&-&-  \\
                DMCNet \cite{DMCNet} & TNNLS'22 &-&-&83.9&59.3&-&-&-&-&-  \\
                MFGNet \cite{MFGNet} & TMM'22 &88.9&70.7&78.3&53.5&-&-&-&-&3.0  \\
                APFNet \cite{APFNet} & AAAI'22 &90.5&73.7&82.7&57.9&-&-&-&-&1.9  \\

                HMFT \cite{HMFT-VTUAV} & CVPR'22 &91.2&74.9&78.8&56.8&75.8&62.7&41.4&35.5&30.2  \\
                HMFT$\_$LT \cite{HMFT-VTUAV} & CVPR'22 &-&-&-&-&-&-&53.6&46.1&8.1  \\
                
                MIRNet \cite{MIRNet} & ICME'23 &90.9&74.4&81.6&58.9&-&-&-&-&30.0  \\
                ECMD \cite{ECMD} & CVPR'23 &90.7&73.5&84.4&60.1&-&-&-&-&18.0  \\
                ViPT \cite{ViPT} & CVPR'23 &-&-&83.5&61.7&-&-&-&-&-  \\
                TBSI \cite{TBSI} & CVPR'23 &-&-&87.1&63.7&-&-&-&-&36.2  \\
                
			\midrule
			USTrack (Ours) &-& \textbf{93.4}& \textbf{78.3} & \textbf{87.4} & \textbf{65.8} & \textbf{86.9} & \textbf{74.4} & \textbf{64.9} & \textbf{55.8} & \textbf{84.2} \\
			\bottomrule
		\end{tabular}
            \caption{\small Comparison with state-of-the-art methods on GTOT, RGBT234, VTUAV short-term subset and long-term subset.}\label{tab1}
            \vspace{-0.6cm}
	\end{center}
\end{table*}

\noindent\textbf{Evaluation on RGBT234 Dataset.}
RGBT234 is currently the most widely used large-scale RGB-T tracking benchmark dataset, consisting of 234 highly aligned image pairs with about 234K image pairs in total. As shown in Tab.\ref{tab1}, compared to the most advanced tracker TBSI, our scores on the metrics PR/SR increased by 0.3$\%$/ 2.1$\%$, and our speed is twice that of TBSI. Both performance and efficiency can prove the effectiveness and efficiency of our method.

\noindent\textbf{Evaluation on GTOT Dataset.}
GTOT is the first standard dataset in the field of RGB-T tracking. It contains 50 RGB-T video sequences and 7 challenge attributes. We also conducted testing on this dataset and achieved SOTA performance. The test results are shown in Tab.~\ref{tab1}. compared with the SOTA methods CMPP and HMFT, our PR/SR scores improved by 0.8$\%$/4.5$\%$ and 2.2$\%$/3.4$\%$ respectively, while maintaining the fastest inference speed.

\subsection{Ablation Experiment and Analysis}

\noindent\textbf{Ablation of Dual Embedding Layer.}
To verify the effectiveness of the dual embedding layer structure, we conducted ablation experiments on the RGBT234 dataset. As a comparison, we have all inputs use the same embedding layer. The results of the ablation experiment are shown in Tab.~\ref{tab2}. The single embedded layer structure resulted in a performance decrease of 1.8$\%$ and 2.6$\%$ in PR and SR scores. The results show that the use of two independent embedding layers can map the features of two modalities into the latent space conducive to fusion, which can alleviate the impact of the intrinsic heterogeneity of modalities on feature fusion based on attention weight.

\begin{table}[htbp]\small
	\begin{center}
		\begin{tabular}{c|c|c}
			\toprule
			Method & PR & SR \\
			\midrule
			Single embedding layer & 85.6 & 63.2 \\
			Dual embedding layer (Ours) & \textbf{87.4} & \textbf{65.8}\\
			\bottomrule
		\end{tabular}
            \caption{ \small Results of the ablation of dual embedding layer.}\label{tab2}
	\end{center}
 \vspace{-0.4cm}
\end{table}


\begin{table}[htbp]\small
	\begin{center}
		\begin{tabular}{c|c|c}
			\toprule
                Method & PR & SR \\
			\midrule
			RGB search region & 86.2 & 64.2 \\
			Thermal search region &86.3 & 64.7\\
			Concatenated region &86.8 &64.2 \\
			Dual predictors with selection (Ours) & \textbf{87.4} & \textbf{65.8}\\
			\bottomrule
		\end{tabular}
            \caption{\small Comparison with different prediction head structures.}\label{tab3}
	\end{center}
        \vspace{-0.9cm}
\end{table}

\begin{figure}[t]
	\centering
	\includegraphics[width=1.0\linewidth]{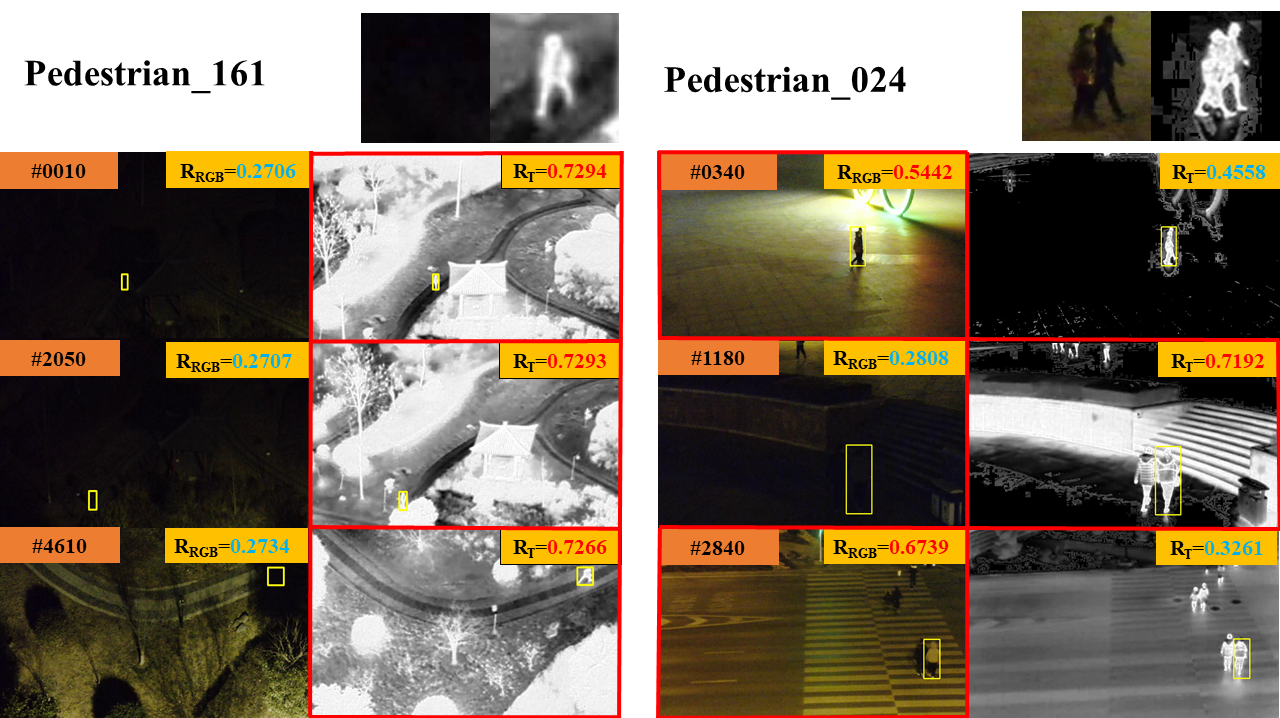}
	\caption{\small The output of the modality reliability evaluation module can intuitively correspond to real scenarios. Best viewed in color.}\label{fig5}
        \vspace{-0.7cm}
\end{figure}

\noindent\textbf{Ablation of Feature Selection Mechanism.}
In order to verify the effectiveness of the feature selection mechanism based on modality reliability, we conducted comparative experiments with several common prediction head structures on RGBT234. We set up single prediction head based on single search region fusion features and single prediction head based on the concatenated fusion features of the two search regions, the experimental results are shown in Tab.~\ref{tab3}. Compared to other prediction heads, our method performs better. As shown in Fig.~\ref{fig5}, we also visualized the actual test sequence, and the visualization results showed that our modality reliability had a good correspondence  with the real scene. USTrack will select search region fusion features with high reliablity scores to output better prediction results.

\section{Conclusion}

In this paper, we propose a highly efficient unified single-stage Transformer RGB-T tracking network USTrack. The core idea of USTrack unifies the three stages of RGB-T tracking network into a single ViT backbone through self-attention mechanism. USTrack can directly extracts fusion features of template and search region under the interaction of modalities, and simultaneously perform the relation modeling between template fusion features and search region fusion features. Additionally, we propose a feature selection mechanism based on modality reliability to achieve better prediction results. Extensive experiments on three popular RGB-T tracking benchmarks demonstrate that our method achieves state-of-the-art performance while maintaining the fastest inference speed 84.2FPS. In particular, evaluation metrics MPR/MSR on the short-term and long-term subsets of the VTUAV increased by 11.1$\%$/11.7$\%$ and 11.3$\%$/9.7$\%$. 

\noindent\textbf{Limitation.} 
USTrack is still a local tracker. In the future, we will explore some efficient local-to-global tracking strategies that are more suitable for RGB-T tracking, further improving the RGB-T tracking performance.

\bibliography{aaai24}

\end{document}